\documentclass[10pt,twocolumn,letterpaper]{article}

\usepackage{cvpr}
\usepackage{times}
\usepackage{epsfig}
\usepackage{graphicx}
\usepackage{amsmath}
\usepackage{amssymb}

\usepackage{color}
\usepackage{float}
\usepackage{amsmath}
\usepackage{amssymb}
\usepackage{algorithm}
\usepackage{verbatim}
\usepackage{caption}
\usepackage{subfigure}
\usepackage{algorithmic}
\usepackage{multirow}
\usepackage{array}

\usepackage[pagebackref=true,breaklinks=true,letterpaper=true,colorlinks,bookmarks=false]{hyperref}

\newcommand{\affmark}[1][*]{\textsuperscript{#1}}

\cvprfinalcopy 

\def\cvprPaperID{1671} 

\ifcvprfinal\fi
\begin{document}

\title{Semi-supervised Transfer Learning for Image Rain Removal\vspace{-4mm}}

\author{Wei Wei\affmark[1,2], Deyu Meng\affmark[1]\thanks{Deyu Meng is the corresponding author.}, Qian Zhao\affmark[1], Zongben Xu\affmark[1], Ying Wu\affmark[2] \\
\affmark[1]School of Mathematics and Statistics, Xi'an Jiaotong University, Xi'an, China\\
\affmark[2]Department of Electrical and Computer Engineering, Northwestern University, IL, USA\\
}


\maketitle
\thispagestyle{empty}

\begin{abstract}
Single image rain removal is a typical inverse problem in computer vision. The deep learning technique has been verified to be effective for this task and achieved state-of-the-art performance. However, previous deep learning methods need to pre-collect a large set of image pairs with/without $\textbf{synthesized}$ rain for training, which tends to make the neural network be biased toward learning the specific patterns of the synthesized rain, while be less able to generalize to $\textbf{real}$ test samples whose rain types differ from those in the training data. To this issue, this paper firstly proposes a semi-supervised learning paradigm toward this task. Different from traditional deep learning methods which only use supervised image pairs with/without synthesized rain, we further put real rainy images, without need of their clean ones, into the network training process. This is realized by elaborately formulating the residual between an input rainy image and its expected network output (clear image without rain) as a specific parametrized rain streaks distribution. The network is therefore trained to adapt real unsupervised diverse rain types through transferring from the supervised synthesized rain, and thus both the short-of-training-sample and bias-to-supervised-sample issues can be evidently alleviated. Experiments on synthetic and real data verify the superiority of our model compared to the state-of-the-arts.
\end{abstract}

\section{Introduction}

Rain streaks and rain drops often occlude or blur the key information of the images captured outdoors. Thus the rain removal task for an image or a video is useful and necessary, which can be served as an important pre-processing step for outdoor visual system. An effective rain removal technique can often help an image/video better deliver more accurate detection or recognition results \cite{kang2012automatic}. 

Current rain removal tasks can be mainly divided into two categories: video rain removal (VRR) and single image rain removal (SIRR). Compared to VRR, which could utilize the temporal correlation among consecutive frames, SIRR is generally much more difficult and challenging without the aid of much prior knowledge capable of being extracted from a single image. Since being firstly proposed by Kang {\em et~al.} \cite{kang2012automatic}, the SIRR problem has been attracting much attention. Recently, deep learning methods \cite{fu2017clearing,fu2017removing,yang2017deep,zhang2017image,derain_zhang_2018,li2018recurrent,fu2018deep,fu2018lightweight,fan2018residual} have been empirically substantiated to achieve state-of-the-art performance for SIRR by training an appropriate, carefully designed network to detect and remove the rain streaks simultaneously.

\begin{figure}[t]

\centering
\includegraphics[height=4.8cm]{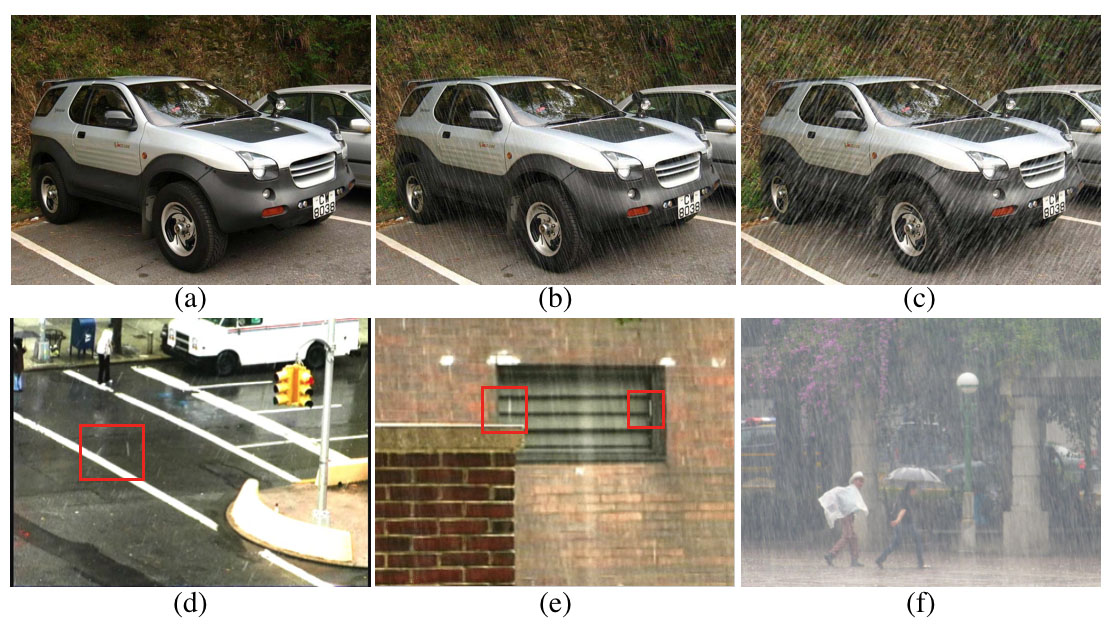}
\vspace{-7mm}
\caption{The comparison of the synthesized rain and real rain. (a) is a clean image; (b), (c) are two synthesized rainy image samples. (d), (e), (f) are real world rainy images.}
\label{Figure1}
\vspace{-3mm}
\end{figure}

Albeit achieving good performance on this task, current deep learning approach still exists some limitations on the methodology. First, for training data, since it is hard to obtain clean/rainy image pairs from real rainy scenarios, previous methods use synthesized data as an alternative, and mainly adopt the strategy of adding the ``fake" rain streaks synthesized by the Photoshop software\footnote{https://www.photoshopessentials.com/photo-effects/rain/} on the clean images. Two samples of such synthesized rainy images are shown in Figure \ref{1}.(b) and (c) (the corresponding clean image is shown in Figure \ref{1}.(a)). Albeit being varied by the rain streak direction and density, the synthesized rainy images still cannot  include sufficiently wider range of rain streak patterns in real rainy images. For instance, in Figure \ref{1}.(d), the rain streaks have multiple directions in a single frame influenced by the wind; in Figure \ref{1}.(e), the rain streaks have multi-layers because of their different distances to the camera; in Figure \ref{1}.(f), the rain streaks produce the effect of aggregation which is similar to fog or mist. Therefore there exists obvious bias between synthetic training data and real testing data in this task, naturally leading to an issue that the network trained on the synthetic training data possibly not capable of being finely generalized to the real test data. 


Meanwhile, one of the main problems for deep learning methods lies on the preliminary conditions that they generally need sufficiently large number of supervised samples (ideal cases are natural images with/without real rain for our task), which are generally time-consuming and cumbersome to collect, in order to train a derain network. However, one generally can easily attain large amount of practical unsupervised samples, {\em i.e.}, real rainy images, while without their corresponding clean ones. How to rationally feed these cheap samples into the network training is not only meaningful and necessary for the investigated task, but also possibly inevitable in the next generation of deep learning to fully prompt its capability on unsupervised data for general image restoration tasks.

%

Due to the inconsistence in the distribution of training data and test data, this task can be naturally viewed as a typical domain adaption problem. How to transfer from learning the synthesized rain patterns (training, supervised) to learning real rain patterns (testing, unsupervised) is crucial. To alleviate the aforementioned issue of previous supervised deep learning methods for the SIRR task, instead of from the perspective of manually collecting more appropriate supervised dataset (real rainy images and their corresponding clean ones) to better suit this task, we propose a novel semi-supervised method attempting to effectively feed unsupervised real rainy images into the network training as well, ultimately expecting to transfer from synthesized rain domain to real rain domain. Different from previous supervised deep learning methods by only using synthesized image pairs as network inputs, our method is capable of fully utilizing unsupervised practical rainy images during training in a mathematically sound manner. Specifically, our model allows both the supervised synthetic data and unsupervised real data being fed into the network simultaneously, and the network parameters can be optimized by the combination of least square residuals (for supervised samples) of network output images of supervised inputs and their ground truth labels, and negative log-likelihood (NLL) losses of a specific parametrized rain distribution (for unsupervised samples) measured by the difference of network output images of unsupervised inputs and their original rainy ones. In this manner, both supervised synthetic and unsupervised real samples can be rationally employed in our method for network training.

In summary, the main contributions of the proposed method are:
\begin{itemize}
\vspace{-2mm}
\item To our knowledge, this is the first work that takes notice of domain adaption issue for SIRR task. We are the first to propose a semi-supervised transfer learning  framework for this task. Different from the previous deep learning SIRR methods, our model can fully take use of the unsupervised real rainy images, which can be easily collected in practice, without need of the corresponding clean ones. Such unsupervised samples not only help evidently reduce the time and labor costs of pre-collecting image pairs with/without real rain for network parameters updating, but also alleviate the over-fitting issue of the deep network on limited rain types covered by only supervised training samples through compensating those unsupervised ones containing more general and practical rain characteristics.

\vspace{-2mm}
\item We provide a general methodology for simultaneously utilizing supervised and unsupervised knowledge for image restoration tasks. For supervised one, the traditional least square loss between network output images and their clean ones can be directly employed. For the unsupervised one, we can rationally formulate the residual between the expected output clean images and their original noisy ones through a likelihood term imposed on a parameterized distribution designed based on the domain understanding for residuals (\eg, rain in our study).
\vspace{-2mm}
\item We design an Expectation Maximization algorithm together with a gradient descent strategy to solve the proposed model. The rain distribution parameters and network parameters can be optimized by sequence in each epoch. Experiments implemented on synthesized rainy images and especially real ones show that our model is capable of transferring from learning synthesized to real rain patterns, thus substantiating the superiority of the proposed method compared to the state-of-the-arts.
\end{itemize}

\vspace{-2mm}
The rest of this paper is organized as follows. In Section \ref{sec2} we detailedly review the previous derain methods along a history line. In Section \ref{sec3}  we present our model as well as the optimization algorithms. We show the experimental results in Section \ref{sec4}  and make conclusion in Section \ref{sec5} .

\section{Related work}
\label{sec2}
\subsection{Single image rain removal methods}

The problem of SIRR was firstly proposed by Kang {\em et~al.} \cite{kang2012automatic}. They detected the rain from the high frequency part of an image based on morphological component analysis and dictionary learning. Chen {\em et~al.}'s \cite{chen2014visual} also operated on the high frequency part of the rainy image but they employed a hybrid feature set, including histogram of oriented gradients, depth of field, and Eigen color, in order to distinguish the rain portions from the image and enhance the texture/edge information. After that, Luo {\em et~al.} \cite{luo2015removing} introduced screen blend model and used discriminative sparse coding for rain layer separation, and the model is solved by greedy pursuit algorithm.  Li {\em et~al.}'s \cite{li2016rain} incorporated patch-based Gaussian mixture model to deliver the prior information of image background and rain layer, and trained the model parameters under pre-collected clean and rainy images. Similarly, Zhang {\em et~al.} \cite{zhang2017convolutional} learned a set of generic sparsity-based and low-rank representation-based convolutional filters to represent background and rain streaks, respectively. Gu {\em et~al.} \cite{gu2017joint} combined analysis sparse representation to represent image large-scale structures and synthesis sparse representation to represent image fine-scale textures, including the directional prior and the non-negativeness prior in their JCAS model. More recently, Zhu {\em et~al.} \cite{zhu2017joint} proposed a joint optimization process that alternates between removing rain-streak details from background layer and removing non-streak details from rain layer. Their model is aided by the rain priors, which are narrow directions and self-similarity of rain patches, and the background prior, which is centralized sparse representation. Chang {\em et~al.} \cite{chang2017transformed} transformed a rainy image into a domain where the line pattern appearance has extremely distinct low-rank structure, and proposed a model with compositional directional total variational and low-rank priors, to deal with the rain streaks as line pattern noise and camera noise at the same time.

While these model-based methods are mathematically sound, they mostly suffer from slow speed when testing because they need to solve an optimization problem. Deep learning has an advantage on test speed and has been substantiated to be effective in many computer vision tasks \cite{lecun1998gradient,krizhevsky2012imagenet,he2016deep}, so does in SIRR. Fu {\em et~al.} firstly introduced deep learning technique for this task in \cite{fu2017clearing}. They trained a convolutional neural network (CNN) with three hidden layers on the high frequency domain of the image. Later, they further ameliorated the CNN by introducing deeper hidden layers, batch normalization and negative residual mapping structure, and achieved better effect \cite{fu2017removing}. To better deal with the scenario of heavy rain images (where individual streaks are hardly seen, and thus visually similar to mist or fog). Yang {\em et~al.} \cite{yang2017deep} exploited a contextualized dilated network with a binary map. In their model, a continuous process of rain streak detection, estimation and removal are predicted in a sequential order. Zhang {\em et~al.} \cite{zhang2017image} applied the mechanism of GAN and introduced a perceptual loss function for the consideration of rain removal problem. Afterwards, they developed a density aware multi-stream dense network for joint rain density estimation and de-raining \cite{derain_zhang_2018}. In summary, these methods learn from synthesized rain data and test their learned network in real scenes.

\subsection{Video rain removal methods}

For literature comprehensiveness, we simply list several representative state-of-the-art video rain removal methods. Since the extra inter-frame information is extremely helpful, these methods showed relatively better reconstruction effect than SIRR methods. Early video derain methods \cite{garg2004detection,barnum2010analysis,bossu2011rain,kim2015video} designed many useful techniques to detect potential rain streaks based on their physical characteristics and removed these detected rain by image restoration  algorithms. In recently years, low-rankness \cite{chen2013generalized,ren2017video}, total variation \cite{jiang2017novel}, stochastic distribution priors \cite{wei2017should}, convolutional sparse coding \cite{li2018video}, neural networks \cite{liu2018d3r} have been applied to the task and achieved satisfying results.

Since the SIRR problem is more difficult in real world with less information provided other than a rainy image, to design an effective SIRR regime is also more challenging beyond VRR ones.

\section{Semi-supervised model for SIRR}
\label{sec3}

We show the framework of our model which includes the training data (both supervised and unsupervised samples) and the network loss in Figure \ref{Figure2}. As introduced aforementioned, our model is capable of feeding not only supervised synthesized rainy images but also unsupervised real rainy images into the network training process, in order to transfer from learning synthesized rain patterns to learning real rain patterns. 

\subsection{Model formulation}

\begin{figure}[t]
\vspace{-2mm}
\centering
\includegraphics[height=6.0cm]{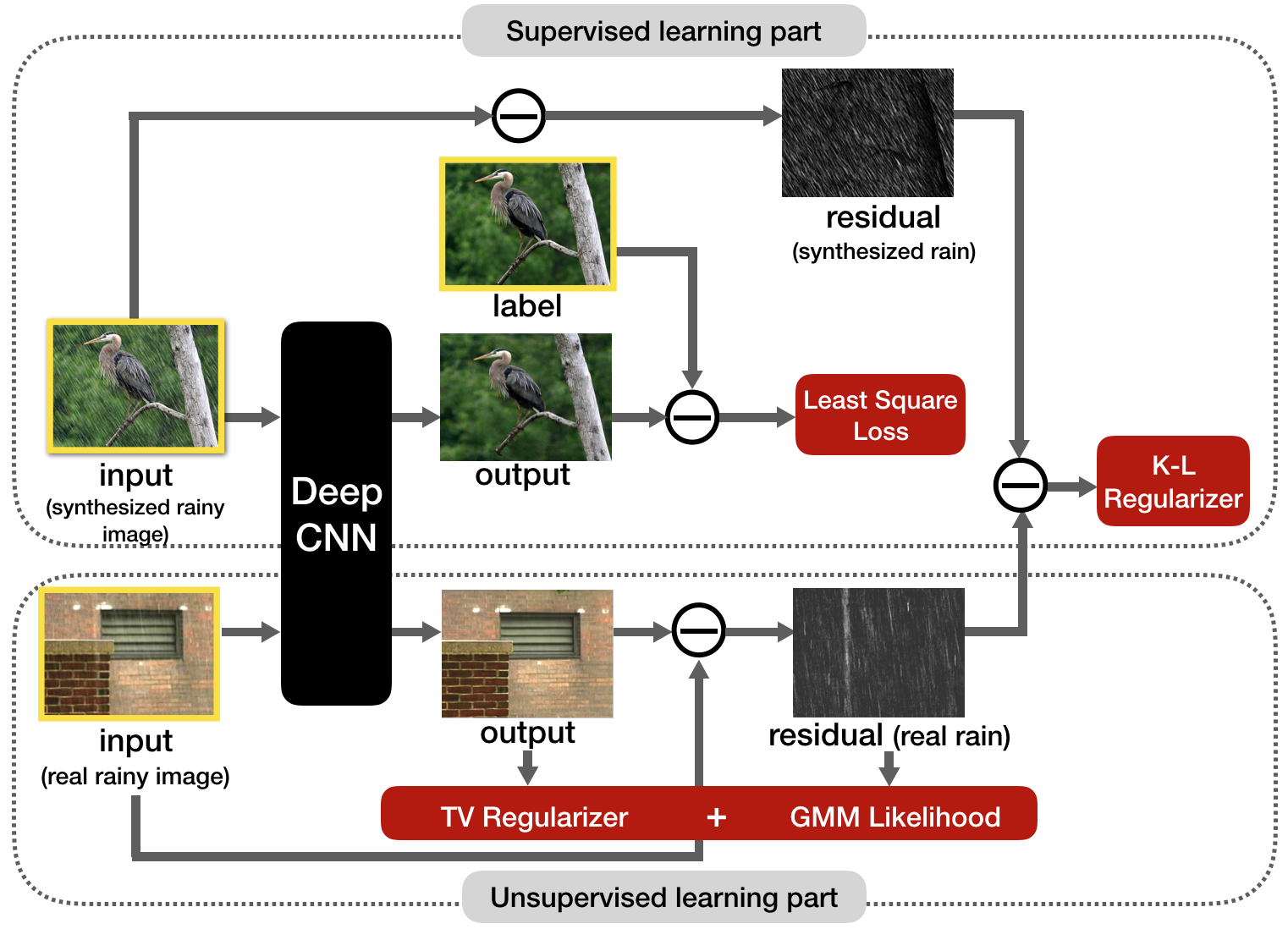}
\vspace{-5mm}
\caption{The flow chart of the proposed method. Those surrounded with concrete yellow square frames are the given inputs for network training. The arrows represent the forward process of network training. The upper panel shows the supervised learning term, which is to minimize the difference of network output and the corresponding clean image using a least square loss. The lower panel shows the unsupervised learning term, which is to minimize an MAP model with a GMM likelihood term imposed on rain distribution and a TV regularization term on background. The network structure and parameters are shared in both parts. A K-L regularizer between the distribution of two types of residuals (extracted rain) is further added  to control the degree of freedom.}
\label{Figure2}
\end{figure}


As shown in Figure \ref{Figure1}.(d,e,f), the real rain usually shows relatively more complex patterns and representations compared to synthesized rain. However, due to the technical defects, these data's ``labels'' ({\em i.e.}, the corresponding clean images) are generally unavailable. Although we can hardly exactly extract the rain layer, as well as the clean background, from a real rainy image, we instead can design a parametrized distribution to finely approximate its stochastic configurations. Since the rain generally contains multi-modal structures due to their occurrence on positions with different distances to the cameras, we can finely approximately express the rain as a Gaussian mixture model (GMM). That is,
\vspace{-1mm}
\begin{equation}
\mathcal{R} \sim \sum_{k=1}^{K} \pi_k\mathcal{N}(\mathcal{R}|\mu_k,\Sigma_k),
\label{1}
\vspace{-2mm}
\end{equation}
where $\pi_k, \mu_k, \Sigma_k$ denote the mixture coefficients, Gaussian distribution means and variances. Mixture models can be universal approximations to any continuous functions if the parameters are learned appropriately, and thus it is suitable to be utilized to describe the rain streaks to-be-extracted from the input rainy image. Thus the negative log likelihood function imposed on these unsupervised samples can be written as:

\vspace{-6mm}
{\small
\begin{equation}
\mathcal{L}_{unsupervised}(\mathcal{R};\Pi, \Sigma) \!=\! -\!\sum_{n=1}^{N} \log \sum_{k=1}^K \pi_k\mathcal{N}(\mathcal{R}_n|0,\Sigma_k),
\label{2}
\end{equation}
\vspace{-4mm}
}

\vspace{-2mm}
\noindent
where $\Pi = \pi_1,...\pi_K$, $\Sigma = \Sigma_1,...\Sigma_K$, $K$ is the number of mixture components, and $N$ is the number of samples. Note that the means of Gaussian distributions are manually set to be zero, and this doesn't affect the results in our experiments.

By utilizing the above encoding manner, we can also construct an objective function for unsupervised rainy images, which can be further used to fine-tune the network parameters through back-propagating its gradients to the network layers.

Meanwhile, we follow the network structure and negative residual mapping skill of DerainNet \cite{fu2017removing} (a deep convolutional neural network) to formulate the loss function on supervised samples. The network which is denoted by $f_{w}(\cdot)$ (here $w$ represents the network parameters) is supposed to remove the rain streaks of input image and output a rain-free one. The classical loss function of CNN is to minimize the least square loss between the expected derain output $f_w(x_i)$ and the ground truth label $y_i$, as shown in the upper panel of Figure \ref{Figure2}. That is, the loss function imposed on the supervised samples is with the following least square form:

\vspace{-3mm}
\begin{equation}
\mathcal{L}_{supervised} = \sum_{i=1}^N ||f_w(x_i)-y_i||_F^2,
\label{4}
\end{equation}

\vspace{-1mm}
\noindent
where $x_i, i =1,...N$ represents the samples of the synthesized rainy image. 

Moreover, since GMM can be adapted to any continuous distribution, in order to let it better fit the real rain samples, we add a constraint that the discrepancy between synthesized rain data domain and real rain data domain is not too far by minimizing a Kullback$-$Leibler divergence between a Gaussian $G_{syn}$ learned from the synthesized rain and the aforementioned mixture of Gaussians $GMM_{real}$ learned from the real rain during training, with a small controlling parameter, as shown in the middle-right of Figure \ref{Figure2}. This is to indicate that our model is expecting to  transfer from synthesized rain to real rain, other than to arbitrary domains. Since this KL divergence is not analytically tractable, we use the minimum of KL divergence  between $G_{syn}$ and each component of $GMM_{real}$ as an empirical and simple substitute, to ensure that at least one component of GMM learned from the real samples is similar to rain. That is,

\vspace{-2mm}
{\small
\begin{equation}
D_{KL}(G_{syn} || GMM_{real}) \simeq \min_k D_{KL}(G_{syn} || GMM^k_{real}),
\label{KL}
\end{equation}}

\vspace{-4mm}
\noindent
where $GMM^k_{real}$ indicates the $k$th component of $GMM_{real}$.

To further remove the potential remained rain streaks in the output image, we add a Total Variation regularizer term to slightly smooth the image. Note that together with the aforementioned likelihood term on rain, a complete MAP model (likelihood + regularizer) is formulated  on the to-be-estimated network outputs of the unsupervised real rain images. It facilitates a right direction for gradient descent to network training on these unsupervised data even without specific explicit guidance of corresponding clean images. 

By combining Eq.(\ref{2}), (\ref{4}), (\ref{KL}) and TV term, the entire objective function to train the network is  formulated as:
\vspace{-5mm}

{\small
\begin{align}
&\mathcal{L}(w,\Pi,\Sigma)  = \sum_{i=1}^{N_1} ||f_w(x_i)-y_i||_F^2  + \alpha\sum_{n=1}^{N_2} ||f_w(\tilde{x})_n||_{TV}   \notag\\
&\!\!\!+ \!\beta D_{KL}(G_x||GMM_{\tilde{x}}) \!-\! \lambda\sum_{n=1}^{N_2} \log \sum_{k=1}^K \pi_k\mathcal{N}(\tilde{x}_n  \!-\!f_w(\tilde{x})_n|0,\Sigma_k),
\label{5}
\end{align}
}

\vspace{-4mm}
\noindent
where $x_i, y_i, i = 1,...N_1$ represent corresponding rainy input and ground truth label sample pairs of the synthesized supervised data, and $\tilde{x}_n, n = 1,...N_2$ represent the rainy input of the real unsupervised data without ground truth labels. Through the last term of Eq. (\ref{5}), the unsupervised data can be fed into the same network with which imposed on the supervised data, and the term $\tilde{x}_n-f_w(\tilde{x})_n$ is the supposed rain extracted from the input rainy image, which is equivalent to $\mathcal{R}_n$ as defined in Eq. (\ref{2}). $\alpha$, $\beta$ and $\lambda$ are the trade-off parameters. Note that when $\alpha, \beta$ and $\lambda$ equal to 0, our model degenerates to the conventional supervised deep learning model \cite{fu2017removing}.

By using such objective setting, the network can be trained not only by the well annotated supervised data, but also purely unsupervised inputs by fully encoding the prior information underlying rain streak distributions. As compared with the traditional deep learning techniques implemented on only supervised samples, the better generalization effect of the network is expected due to the fact that it facilitates a rational transferring effect from the supervised samples to unsupervised types of rain.

\subsection{The EM algorithm}

Since the loss function in Eq. (\ref{5}) is intractable, we use the Expectation Maximization algorithm \cite{dempster1977maximum} to iteratively solve the model. In E step, the posterior distribution which represents the responsibility of certain mixture component is calculated. In M step, the mixture distribution and the convolutional neural network parameters are updated.. 

\noindent
$\mathbf{E~step:}$  Introduce a latent variable $z_{nk}$ where $z_{nk}\in\{0, 1\}$ and $\sum_{k=1}^K z_{nk}=1$, indicating the assignment of noise term $(\tilde{x}_n-f_w(\tilde{x})_n)$ to a certain component of the mixture model. According to the Bayes' theorem, the posterior responsibility of component $k$ for generating the noise is given by:

\vspace{-1mm}
\begin{equation}
\gamma_{nk} = \frac{\pi_k\mathcal{N}(\tilde{x}_n-f_w(\tilde{x})_n|0,\Sigma_k)}{\sum_k \pi_k\mathcal{N}(\tilde{x}_n-f_w(\tilde{x})_n|0,\Sigma_k)}.
\label{6}
\end{equation}

\noindent
$\mathbf{M~step:}$ After the E step, the loss function in Eq. (\ref{5}) is unfolded into a differential one with respect to GMM parameters, shown as:

\vspace{-5mm}
{\small
\begin{align}
&\min_{w,\Pi,\Sigma} ~~  \lambda\sum_{n=1}^{N_2} \sum_{k=1}^K \gamma_{nk}(\frac{(\tilde{x}_n\!-\!f_w(\tilde{x})_n)^2}{2\Sigma_k}\!+\!\frac{1}{2}\!\log|\Sigma_k|\!-\!\log\pi_k) \! \notag\\
\vspace{-2mm}
&\!\!\!+ \!\sum_{i=1}^{N_1} ||f_w(x_i)\!-\!y_i||_F^2 \!+\! \alpha\!\sum_{n=1}^{N_2}
||f_w(\tilde{x}_n)||_{TV} \!+\! \beta D_{KL}(G_x||GMM_{\tilde{x}}).
\label{7}
\end{align}
}

\vspace{-5mm}
\noindent
The closed-form solution of mixture coefficients and Gaussian covariance parameters are \cite{dempster1977maximum}:

\vspace{-4mm}
\begin{equation}
N_{k}=\sum_{n=1}^N \gamma_{nk},\qquad\pi_k=\frac{N_{k}}{N},
\label{8}
\end{equation}
\vspace{-3mm}
\begin{equation}
\Sigma_{k}=\frac{1}{N_k}\sum_{n=1}^N \gamma_{nk}(\tilde{x}_n-f(\tilde{x})_n)^2, k=1,...K.
\label{9}
\end{equation}

\vspace{-1mm}
\noindent
Then we can employ the gradient methods to optimize the objective function as defined in Eq. (\ref{7}) and the gradient so calculated can thus be easily back propagated to the network to gradually ameliorate its parameters $w$. We readily utilize Adam \cite{kingma2014adam}, the off-the-shelf first order gradient optimization algorithm, for network parameter training on the objective function (\ref{7}) imposed on both synthesized supervised and real unsupervised training samples.

\subsection{Discussions on domain transfer learning}
\label{sec3.3}

\begin{figure*}[t]
\vspace{-3mm}
\centering
\includegraphics[height=7.2cm]{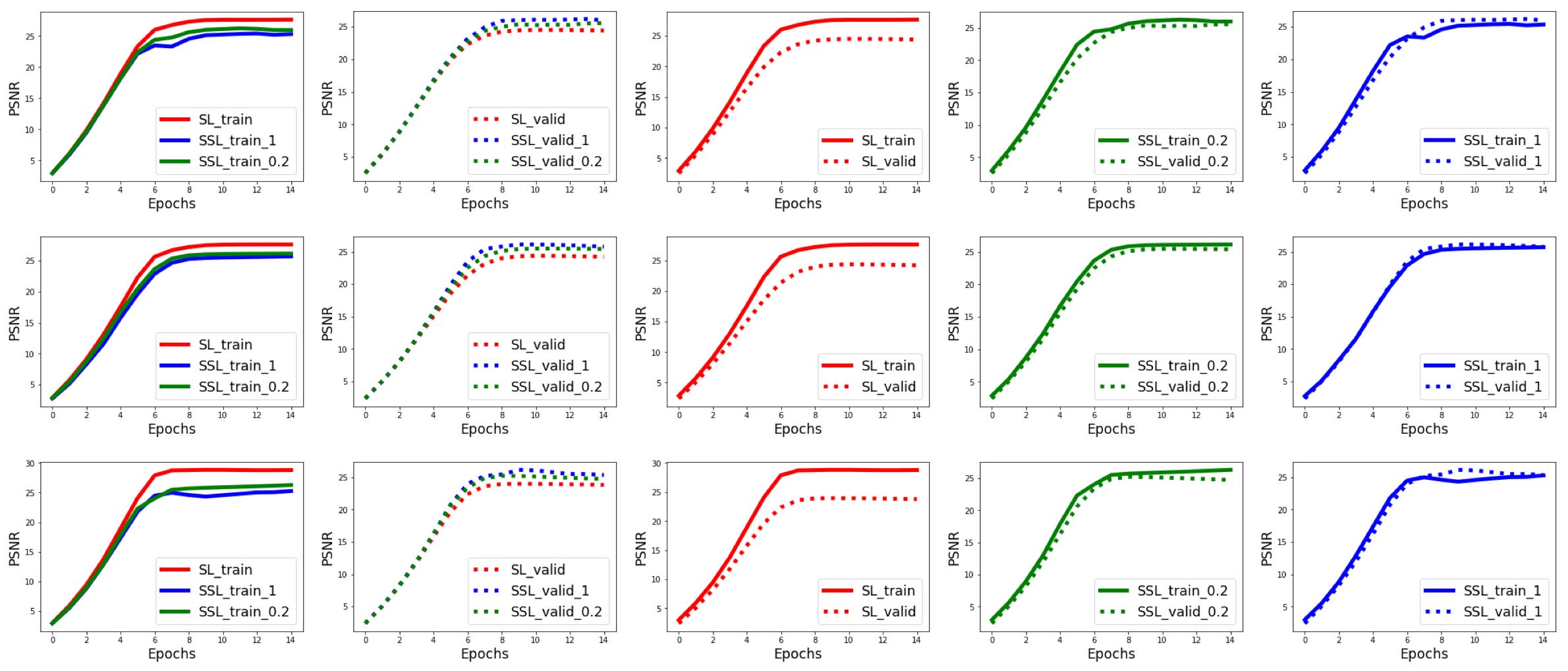}
\vspace{-3mm}
\caption{The PSNR trend graph of supervised training data and validation data during training process. In all subgraphs, the solid line represents trend of supervised training data and the dotted line represents the trend of validation data. Note that they in distinct domain because of the different rain-synthesized way. The red, green, blue lines represent the unsupervised term controlling parameter $\lambda$ in Eq. (\ref{5}) equals to 0 (equivalent to supervised learning), 0.2 and 1, respectively. The three rows use five hundred, five thousand and ten thousand image patches as the training data from top to bottom.}
\label{Figure3}
\end{figure*}

The main difference of the proposed method from the other supervised deep learning SIRR methods is the involvement of the real world rainy images whose ground truth rain-free images or ground-truth rain images) are unavailable during training. One main motivation for this investigation is that the manually synthesized rain shapes usually differ from real ones collected in practice. According to several SIRR methods in the framework of deep learning \cite{fu2017clearing,fu2017removing,yang2017deep,derain_zhang_2018}, 
clean images are used to synthesize rainy images by Photoshop software. Although each clean image is supposed to synthesize several different type of rainy image, as shown in upper panel of Figure \ref{Figure1}, the difference of scale, illuminance and distance to the camera of the real rain streaks and usually accompanied fog or mist visual effect are hardly sufficiently considered, thus yielding nonnegligible gap between the synthesized rainy images for training and the real rainy images for testing.


In our method, the involvement of the unsupervised real rainy data alleviates this problem. As shown in Figure \ref{Figure3}, we use the same synthesized rainy data with \cite{fu2017removing} as the supervised training data. To empirically show the domain transfer capability and verify the superiority of our model on this point, we use a different way to synthesize rainy images introduced in \cite{wei2017should}, and separate them as unsupervised input set and validation set. Therefore the supervised training rain and validation rain lie in distinct domains. We found that our model shows better capability to overcome the gap and transfer from the training data domain to validation data domain. Although our semi-supervised model not extremely finely fit the effect of the training data when the unsupervised term in our loss function Eq. (\ref{5}) plays more important role (as shown in Column 1 of Figure \ref{Figure3}, green and blue lines are our semi-supervised model, with different unsupervised term parameters), Column 2 of Figure \ref{Figure3} reflects that our model has better effect on the target domain (solid line represents supervised data domain while dotted line represents target domain). Moreover, with the training dataset booming, the baseline supervised CNN (red line in Figure \ref{Figure3}) tends more and more to achieve specific patterns of the training data (\ie, the performance of training data improve), thus less being  generalized to the validation data (\ie, the performance of testing data does not improve correspondingly, even slightly worsen) if they lie in separate domain, as shown in Column 3 of Figure \ref{Figure3}. However, the involvement of the unsupervised term in our loss function can effectively alleviate this issues, as shown in Column 4 and 5 of Figure \ref{Figure3}, which is critical in real rain removal task.

\section{Experimental results}
\label{sec4}

In this section, we evaluate our methods both on synthesized rainy data and real world rainy data. The compared methods include the discriminative sparse coding based method (DSC) \cite{luo2015removing}, layer priors based method (LP) \cite{li2016rain}, CNN method \cite{fu2017removing}, joint bi-layer optimization (JBO) \cite{zhu2017joint}, multi-task deep learning method (JORDER) \cite{yang2017deep} and multi-stream dense net (DID-MDN) 
\cite{derain_zhang_2018}. These methods include conventional unsupervised model-driven methods and more recent supervised data-driven deep learning methods. Our method to some extent can be viewed as an intrinsic combination of both methodologies. 

\subsection{Implementation details}

For supervised training data, we use one million 64$\times$64 synthesized rainy/clean image patch pairs which are the same with the baseline CNN method \cite{fu2017removing}. For unsupervised training data, we collect the real world rainy images from the dataset provided by \cite{yang2017deep,wei2017should,zhang2017image} and Google image search. We randomly cropped one million 64$\times$64 image patches from these images to constitute the unsupervised samples. Batch size is 20. The initial learning rate is $10^{-3}$, decaying by multiplying 0.1 after every 5 epochs. We train 15 epochs in total.  The training is implemented using Tensorflow \cite{abadi2016tensorflow}.

We design the number of GMM components as 3. For the trade-off parameter $\lambda$, we simply set it as 0.5 throughout all our experiments. The parameter $\alpha$ which controls the TV smoothing term is set as a small value $10^{-5}$. The parameter $\beta$ which controls the KL divergence term is set to be $10^{-9}$. The network structures and related parameters are directly inherited from the baseline method \cite{fu2017removing}.

\begin{figure}[t]
\centering
\vspace{-2mm}
\subfigure[Input] {\includegraphics[width=0.22\textwidth]{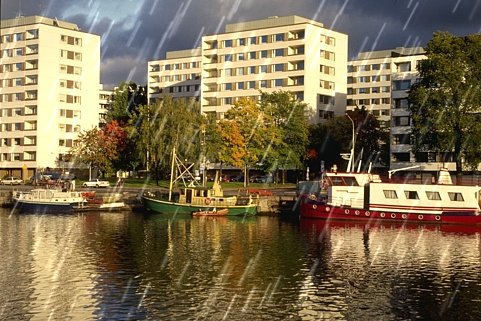}}
\subfigure[Ground truth] {\includegraphics[width=0.22\textwidth]{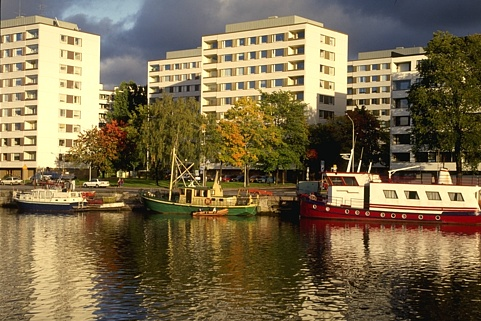}}
\vspace{-2mm}
\subfigure[DSC \cite{luo2015removing}] {\includegraphics[width=0.22\textwidth]{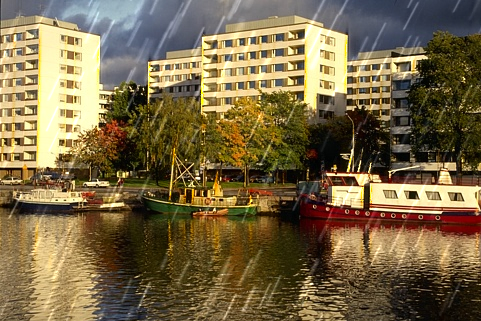}}
\subfigure[LP \cite{li2016rain}] {\includegraphics[width=0.22\textwidth]{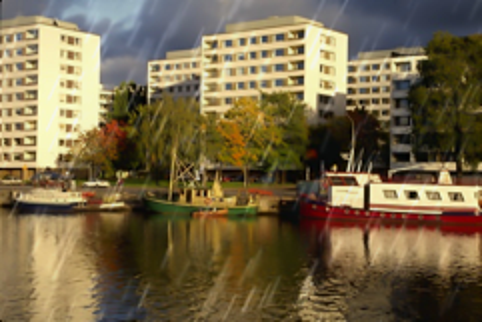}}
\vspace{-2mm}
\subfigure[CNN \cite{fu2017removing}] {\includegraphics[width=0.22\textwidth]{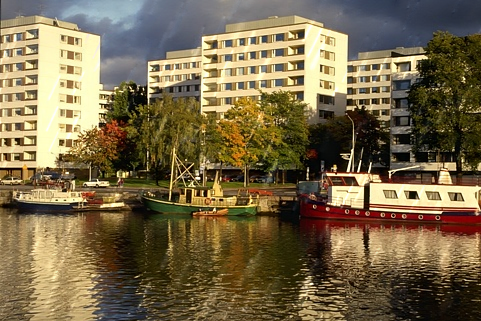}}
\subfigure[JORDER \cite{yang2017deep}] {\includegraphics[width=0.22\textwidth]{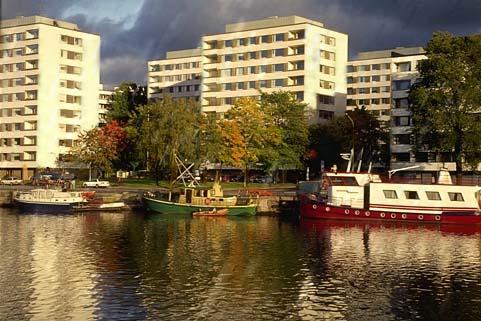}}
\vspace{-2mm}
\subfigure[JBO \cite{zhu2017joint}] {\includegraphics[width=0.22\textwidth]{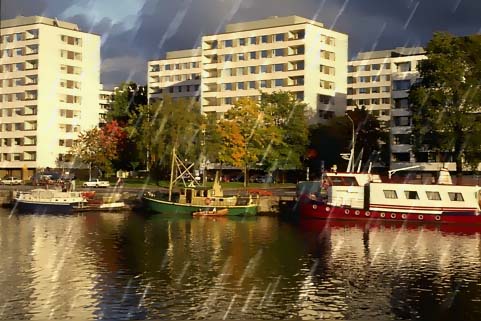}}
\subfigure[Ours] {\includegraphics[width=0.22\textwidth]{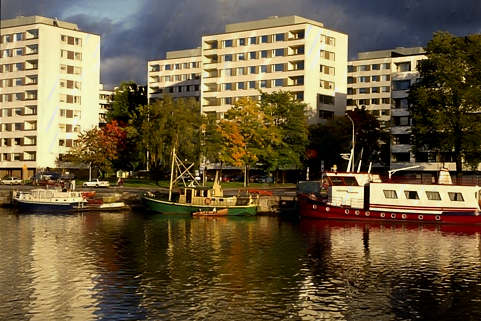}}
\vspace{-2mm}
\caption{Synthesized rain removal results under the sparse rain streaks scenario.}\label{sparse}
\vspace{-2mm}
\end{figure}

\begin{figure}[t]
\centering
\vspace{-2mm}
\subfigure[Input] {\includegraphics[width=0.22\textwidth]{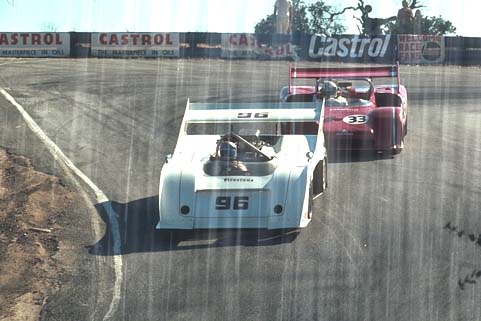}}
\subfigure[Ground truth] {\includegraphics[width=0.22\textwidth]{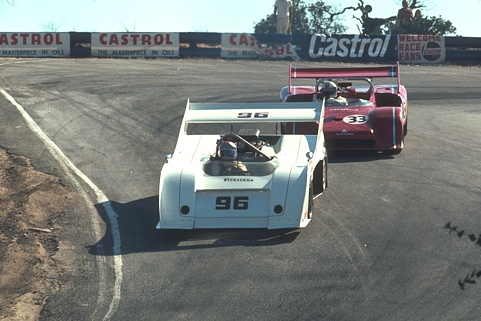}}
\vspace{-2mm}
\subfigure[DSC \cite{luo2015removing}] {\includegraphics[width=0.22\textwidth]{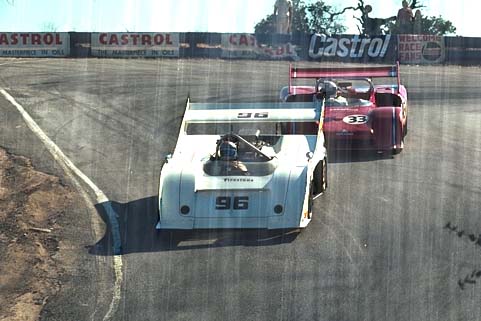}}
\subfigure[LP \cite{li2016rain}] {\includegraphics[width=0.22\textwidth]{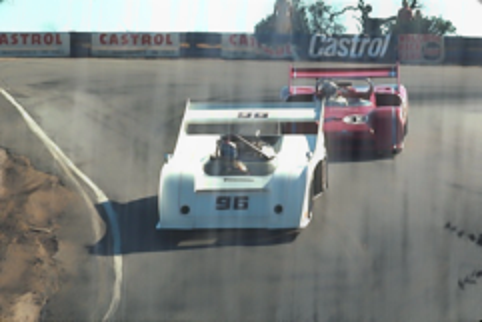}}
\vspace{-2mm}
\subfigure[CNN \cite{fu2017removing}] {\includegraphics[width=0.22\textwidth]{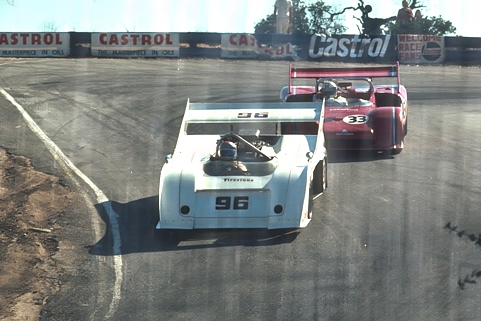}}
\subfigure[JORDER \cite{yang2017deep}] {\includegraphics[width=0.22\textwidth]{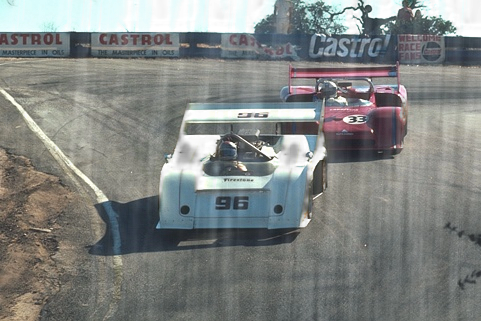}}
\vspace{-2mm}
\subfigure[JBO \cite{zhu2017joint}] {\includegraphics[width=0.22\textwidth]{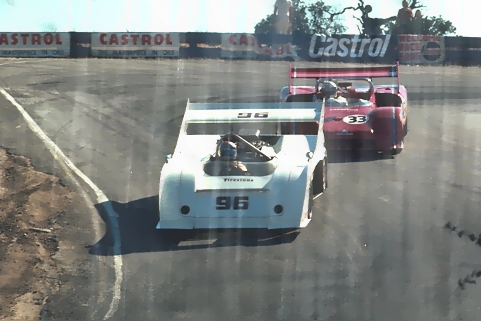}}
\subfigure[Ours] {\includegraphics[width=0.22\textwidth]{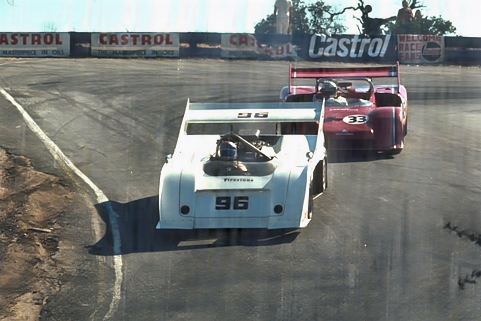}}
\vspace{-2mm}
\caption{Synthesized rain removal results under the dense rain streaks scenario.}\label{dense}
\vspace{-2mm}
\end{figure}

\subsection{Experiments on synthetic images}


\setlength{\tabcolsep}{4pt}
\begin{table*}[!h]
\begin{center}
\caption{Mean PSNR comparison of two groups of data on synthesized rainy images.}
\vspace{-2mm}
\begin{tabular}{|c|c|c|c|c|c|c|c|c|}
\hline
Dataset & Input & DSC\cite{luo2015removing} & LP\cite{li2016rain} & JORDER\cite{yang2017deep} & CNN\cite{fu2017removing} & JBO\cite{zhu2017joint} &DID-MDN\cite{derain_zhang_2018}& Ours \\
\hline
Dense & 17.95 & 19.00 & 19.27 & 18.75 & 19.90 & 18.87 & 18.60& {\bf 21.60} \\
Sparse & 24.14 & 25.05 & 25.67 & 24.22 & 26.88 & 25.24 & 25.66& {\bf 26.98} \\
\hline
\end{tabular}\vspace{-8mm}
\label{table}
\end{center}
\end{table*}
\setlength{\tabcolsep}{1.4pt}



In this subsection, we evaluate the rain removal effect of our method with synthetic data by both visual quality and performance metric. We use the skill of \cite{wei2017should} to synthesize the rainy image as test data. Considering the complexity and multiformity of the rain streaks, we compare our methods with others under two different scenarios: sparse rain streaks and dense rain streaks. In each scenario we use ten test images. Figure \ref{sparse} shows an example of synthetic data with sparse rain streaks. The added rain streaks are sparse but with multiple lengths and layers, in consideration of the different distance to the camera. As shown in Figure \ref{sparse}, the DSC method \cite{luo2015removing} and JBO method \cite{zhu2017joint} fail to remove the main component of the rain streaks. The LP method \cite{li2016rain} tends to blur the visual effect of the image and over-smooth the texture and edge information. The two deep learning methods CNN \cite{fu2017removing} and JORDER \cite{yang2017deep} have better rain removal effects, but rain streaks sill clearly exist in their results. Comparatively, our method could better remove the sparse rain streaks and keep the background information.

We also design the experiments with dense rain streaks scenario. In real world, the dense rain streaks have the effect of aggregation, blurring the image similar to fog or mist when the rain is heavy. In Figure \ref{dense}, the added rain is heavy, with not only the long rain streaks, but also the brought blurring effect damaging the image visual quality. As shown in Figure \ref{dense}, the results of DSC \cite{luo2015removing}, JORDER \cite{yang2017deep} and JBO \cite{zhu2017joint} still have obvious rain streaks, while LP \cite{li2016rain} still over-smoothes the image. Compared with the baseline CNN method \cite{fu2017removing}, our method has better restoration results.


Since the ground truth is known for the synthetic experiments, we use the most extensive performance metric Peak Signal-to-Noise Ratio (PSNR) for a quantitative evaluation. As is evident in Table \ref{table}, our method attains the best PSNR in both two groups of data with different scenarios, in agreement with the visual effect in Figures \ref{sparse} and \ref{dense}.

\begin{figure*}[!t]
\vspace{-4mm}
\centering 
\subfigure{\includegraphics[width=0.16\textwidth]{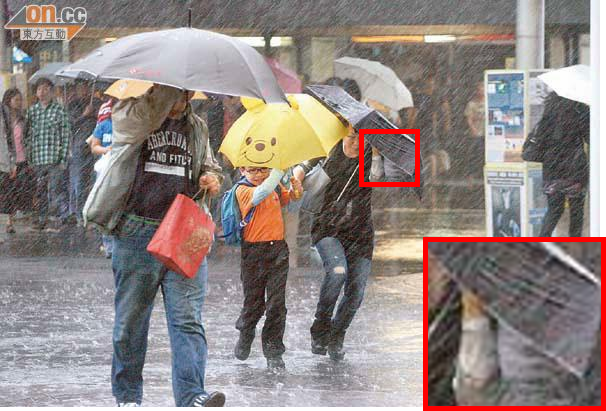}}
\subfigure{\includegraphics[width=0.16\textwidth]{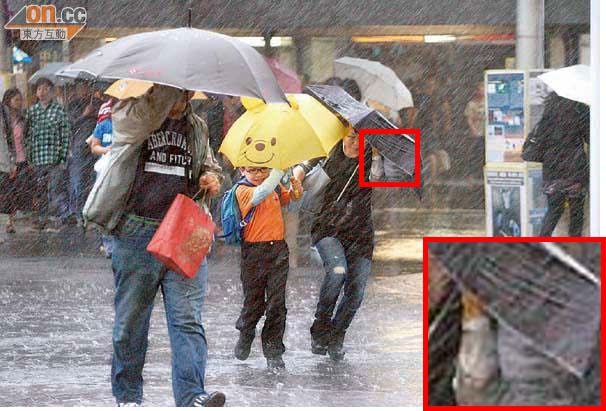}}
\subfigure{\includegraphics[width=0.16\textwidth]{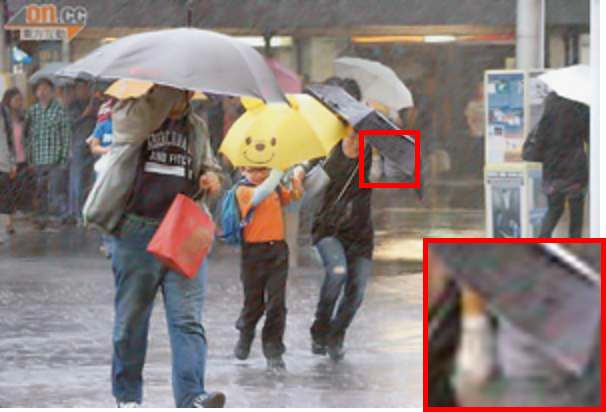}}
\subfigure{\includegraphics[width=0.16\textwidth]{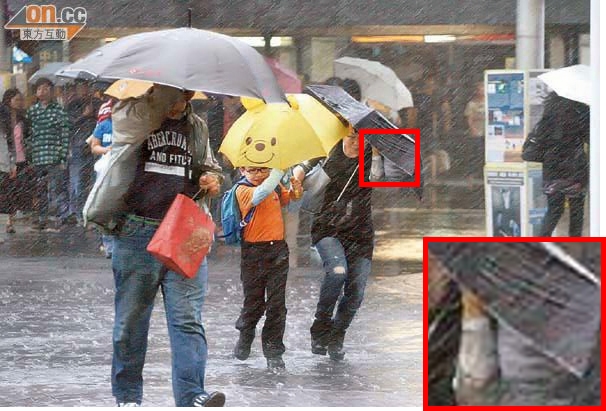}}
\subfigure{\includegraphics[width=0.16\textwidth]{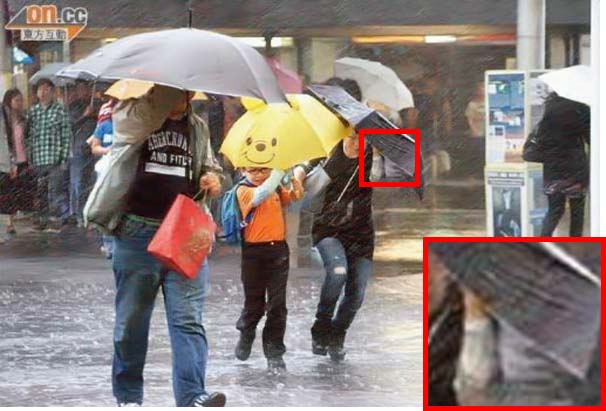}}
\subfigure{\includegraphics[width=0.16\textwidth]{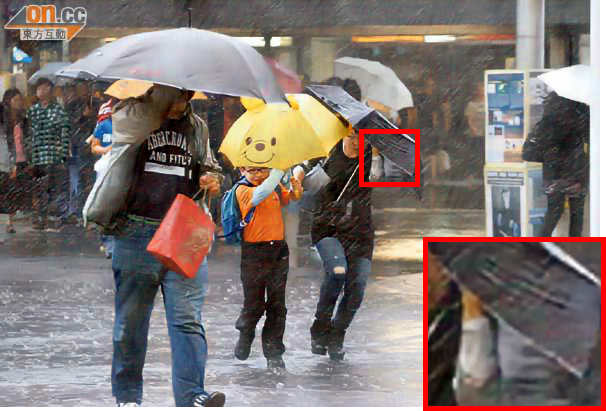}}

\subfigure{\includegraphics[width=0.16\textwidth]{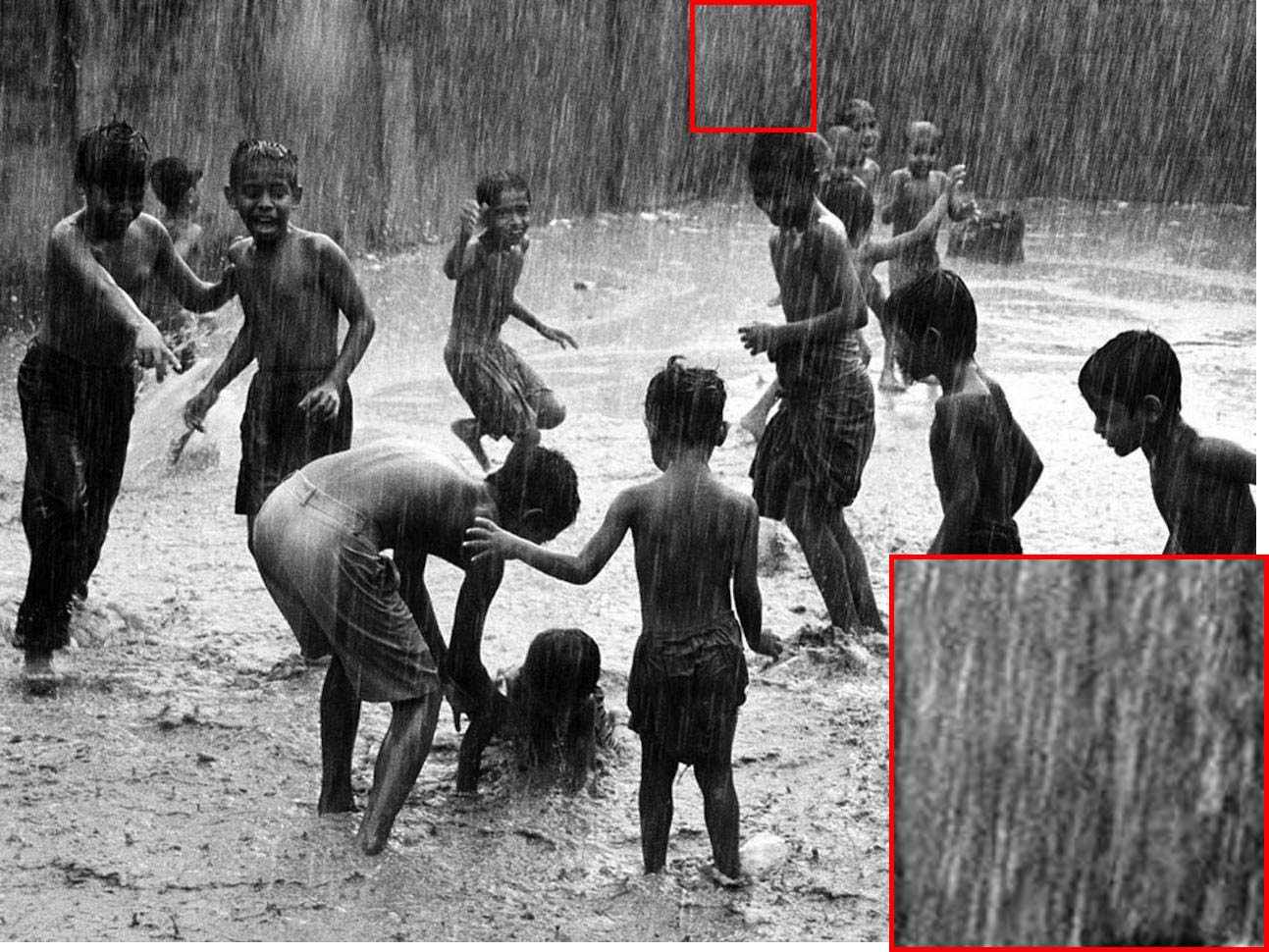}}
\subfigure{\includegraphics[width=0.16\textwidth]{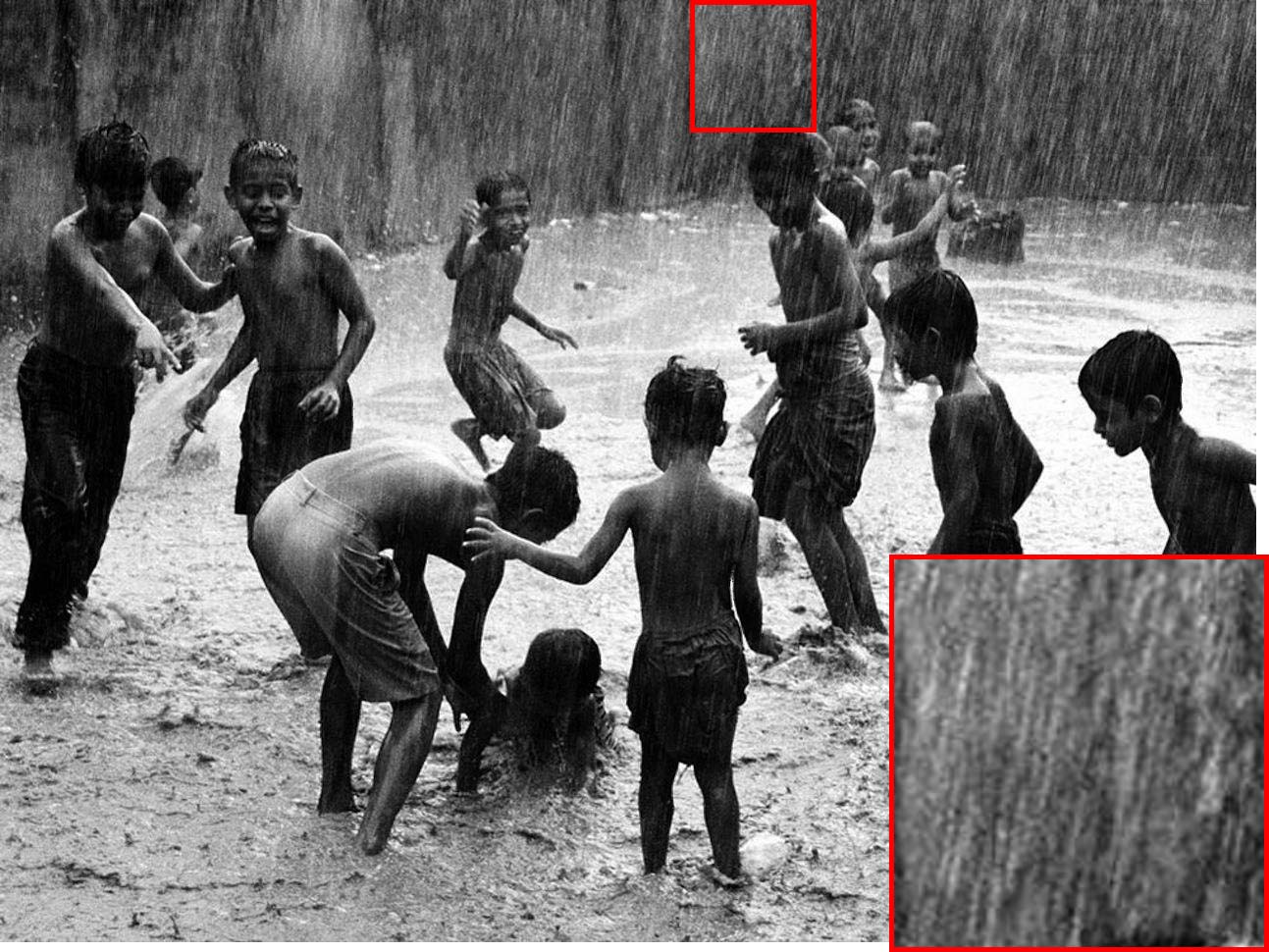}}
\subfigure{\includegraphics[width=0.16\textwidth]{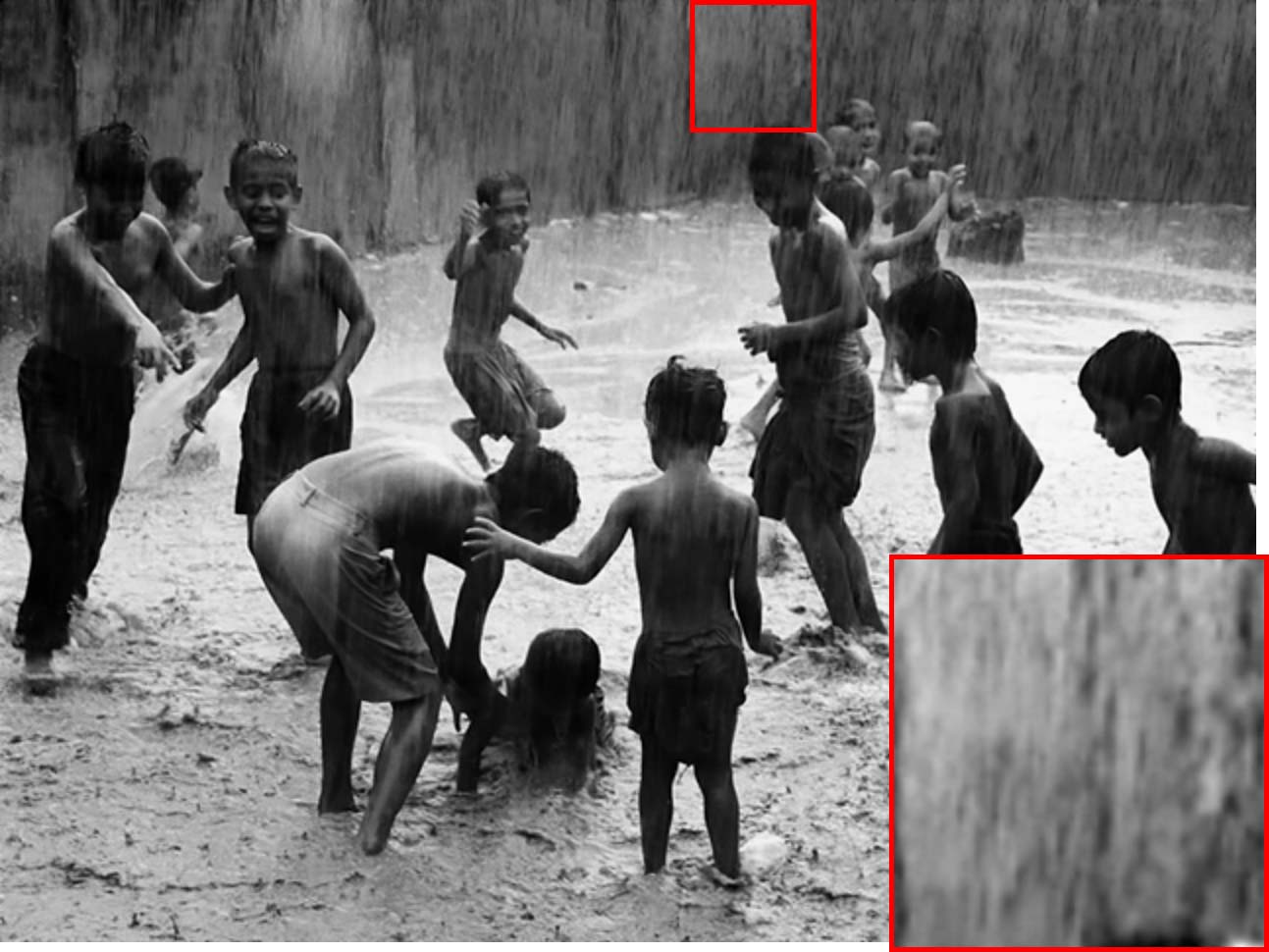}}
\subfigure{\includegraphics[width=0.16\textwidth]{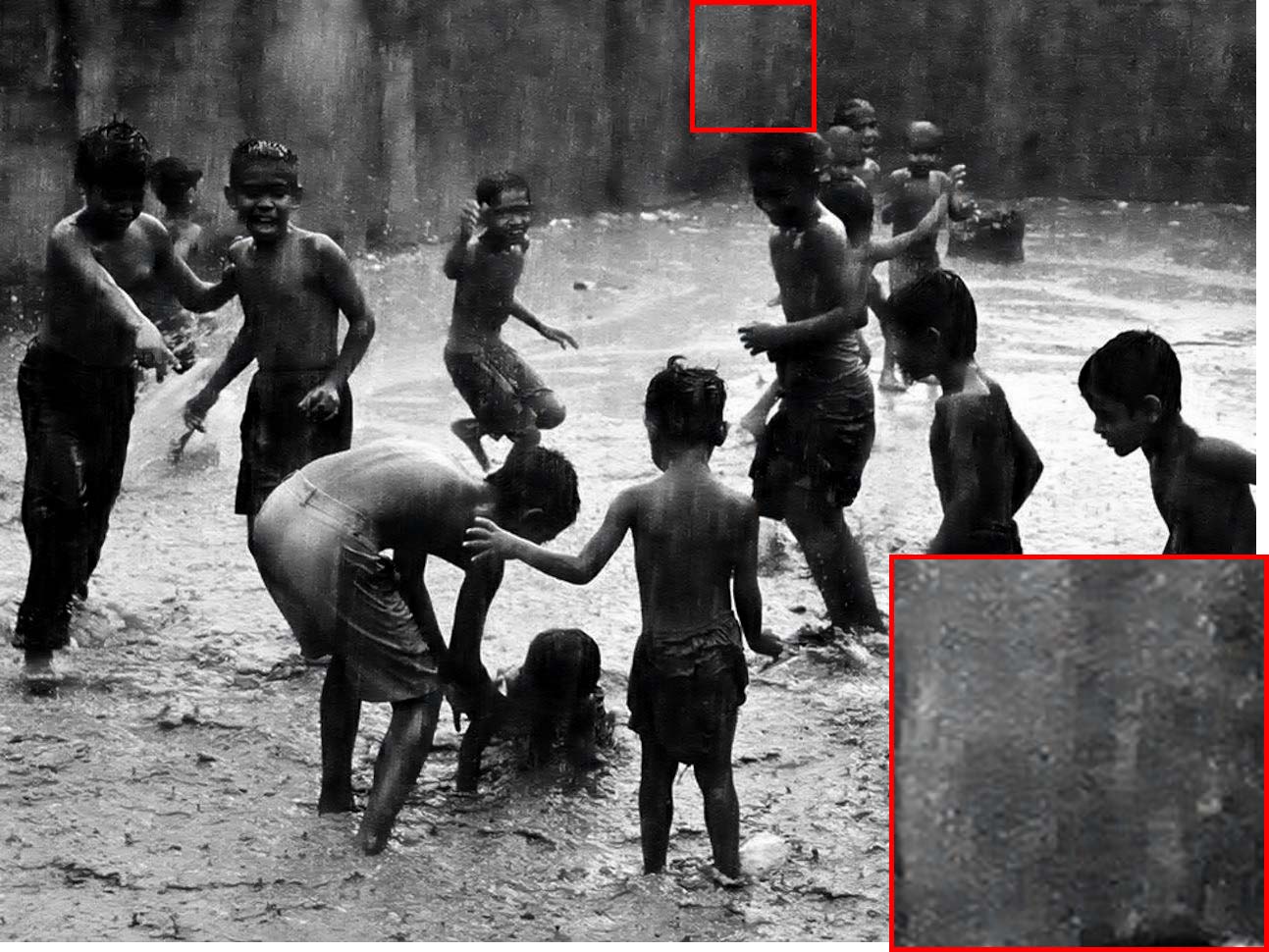}}
\subfigure{\includegraphics[width=0.16\textwidth]{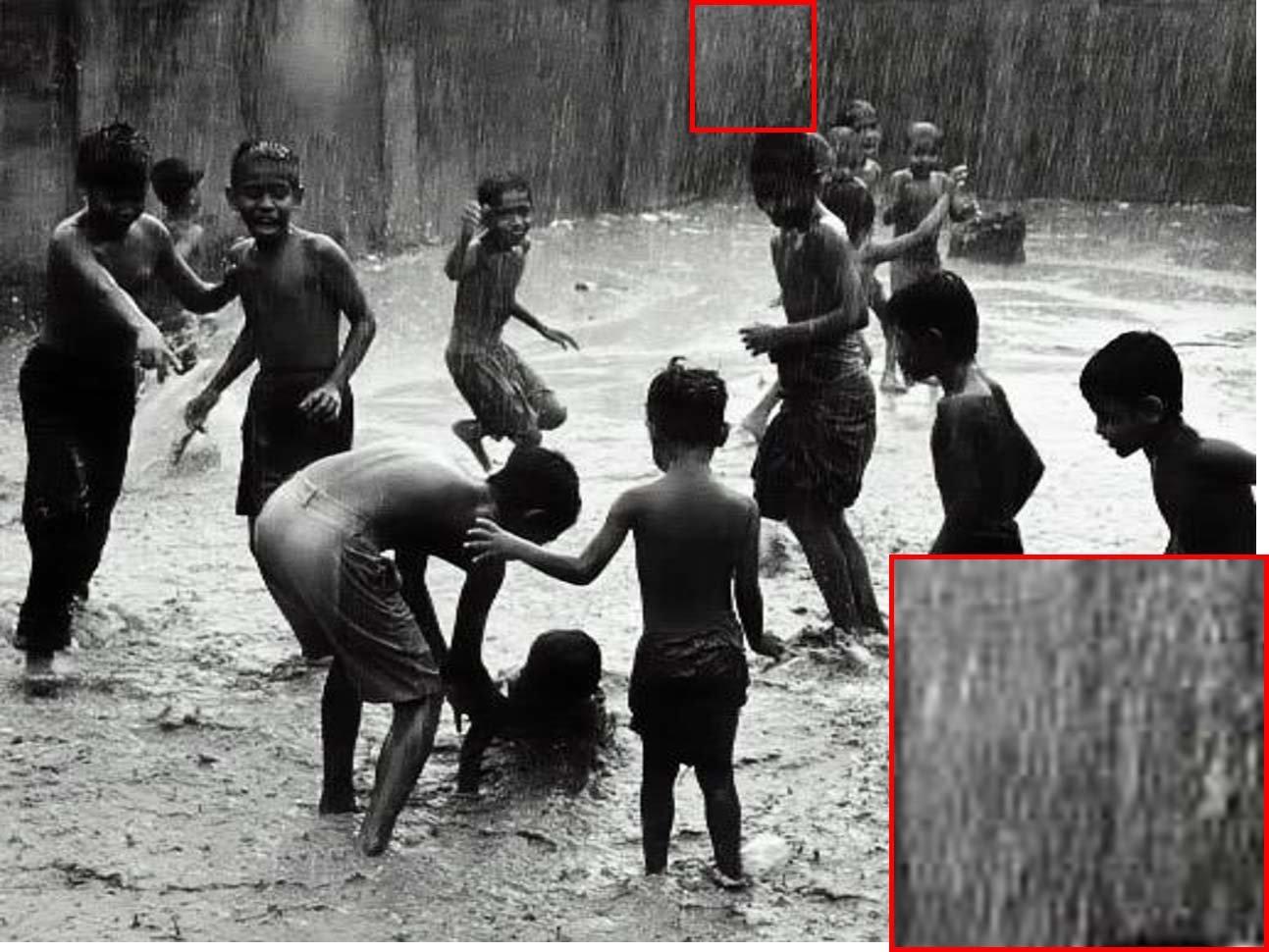}}
\subfigure{\includegraphics[width=0.16\textwidth]{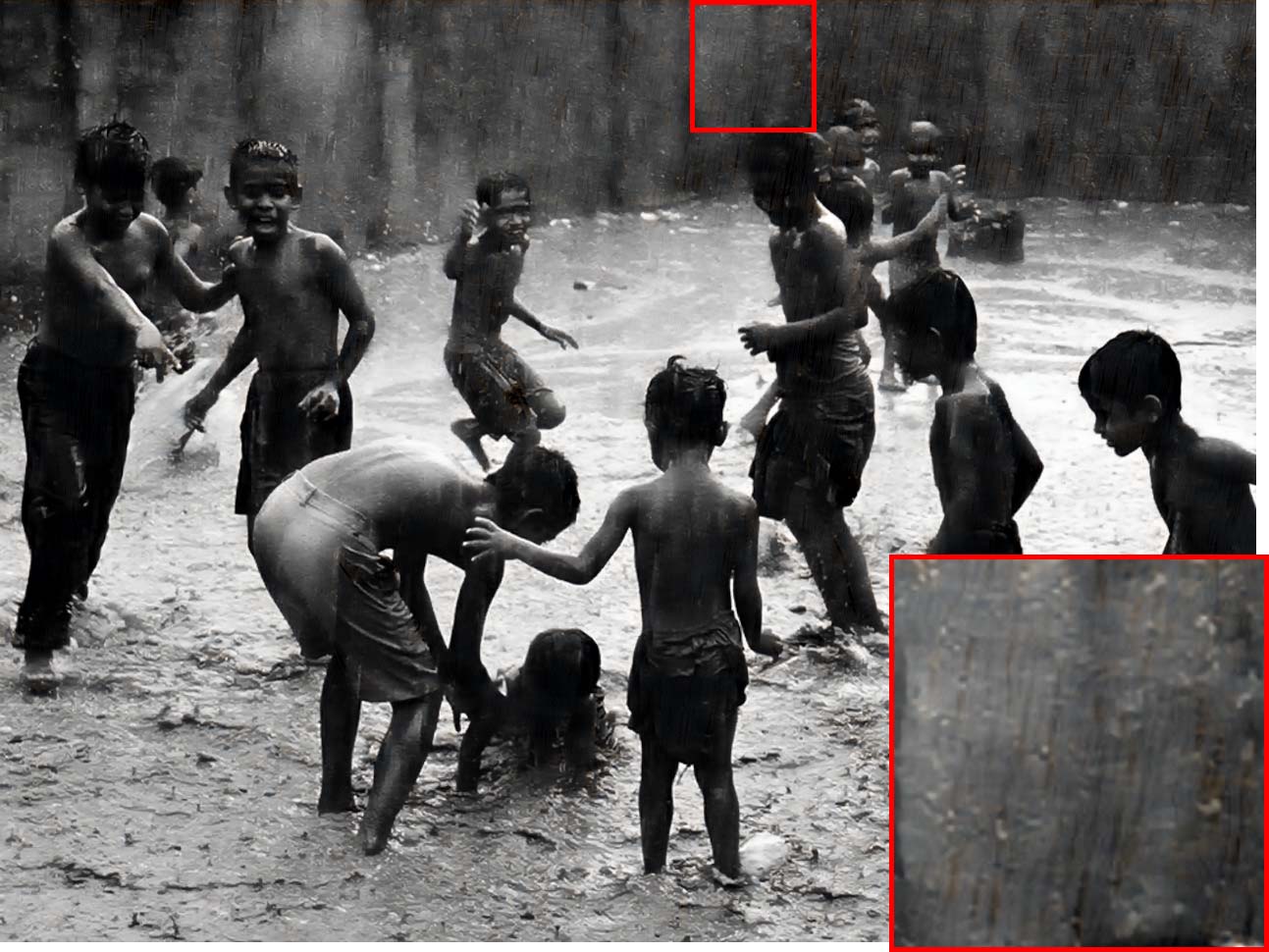}}

\subfigure{\includegraphics[width=0.16\textwidth]{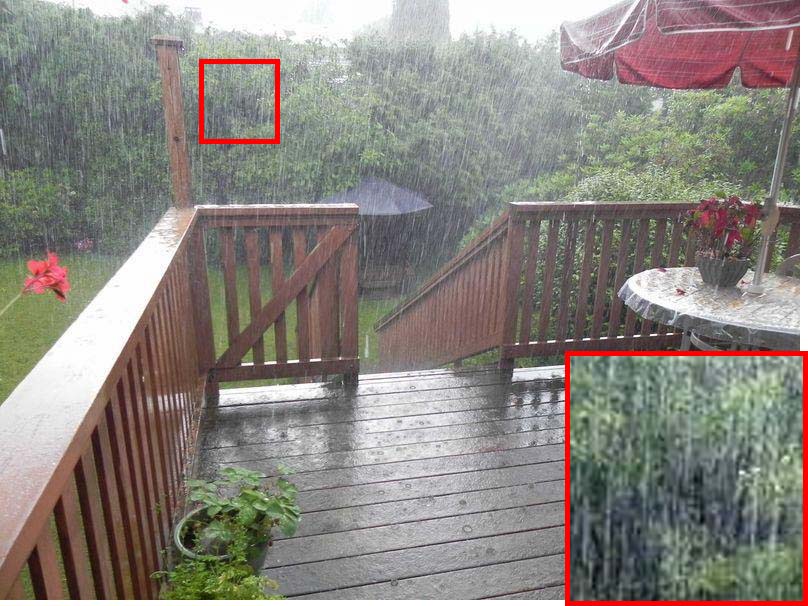}}
\subfigure{\includegraphics[width=0.16\textwidth]{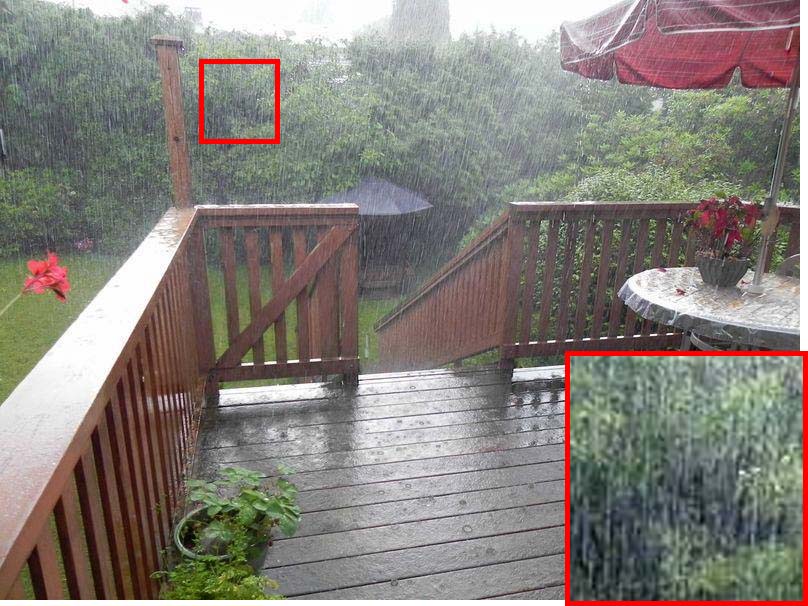}}
\subfigure{\includegraphics[width=0.16\textwidth]{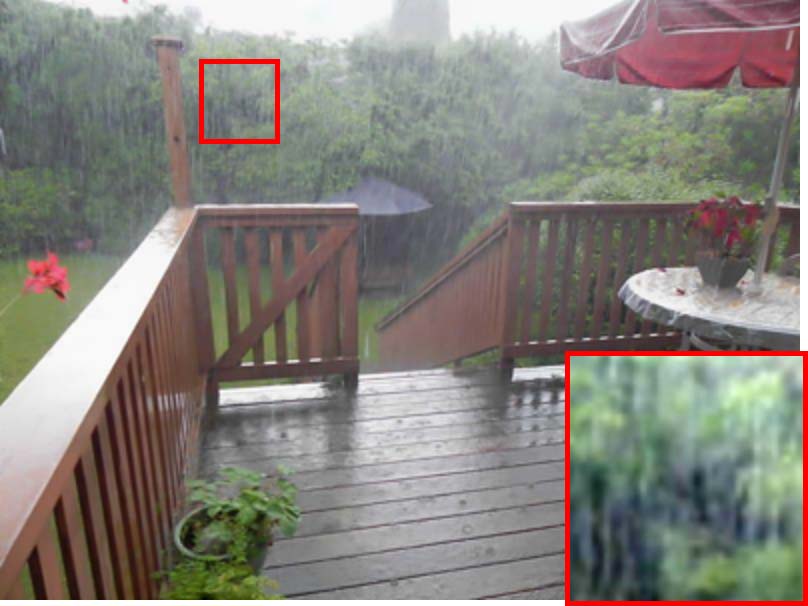}}
\subfigure{\includegraphics[width=0.16\textwidth]{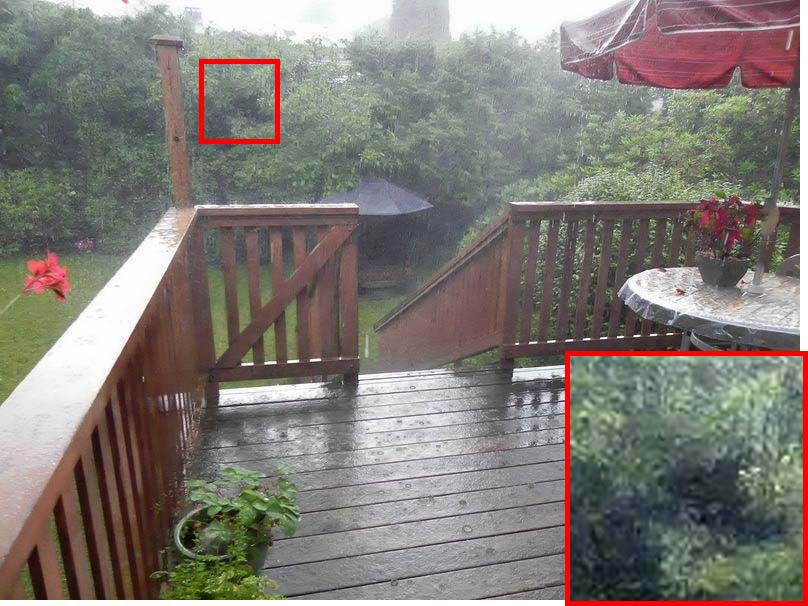}}
\subfigure{\includegraphics[width=0.16\textwidth]{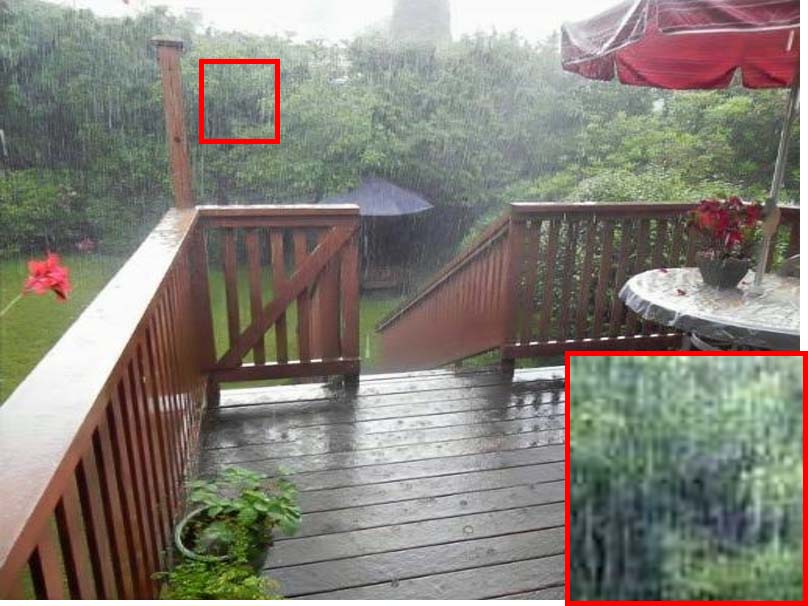}}
\subfigure{\includegraphics[width=0.16\textwidth]{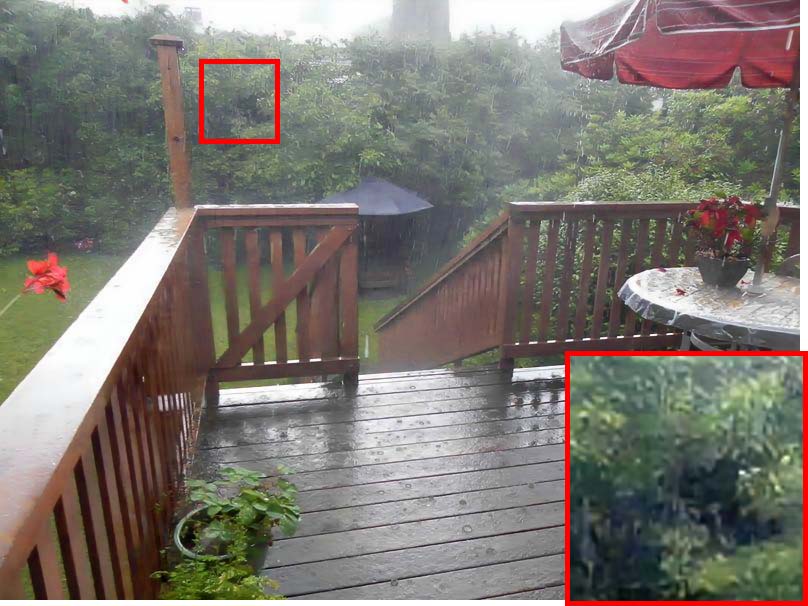}}


\subfigure{\includegraphics[width=0.16\textwidth]{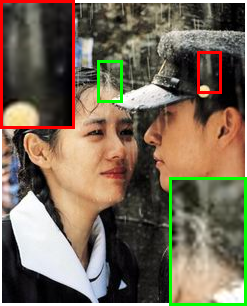}}
\subfigure]{\includegraphics[width=0.16\textwidth]{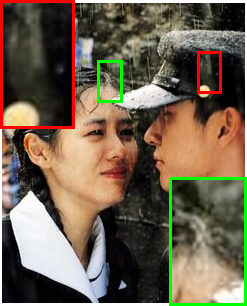}}
\subfigure{\includegraphics[width=0.16\textwidth]{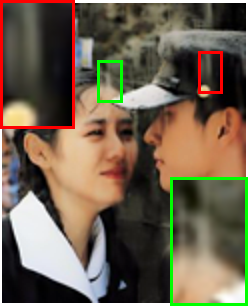}}
\subfigure{\includegraphics[width=0.16\textwidth]{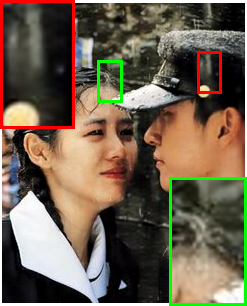}}
\subfigure{\includegraphics[width=0.16\textwidth]{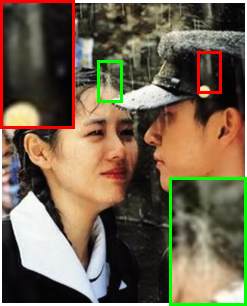}}
\subfigure{\includegraphics[width=0.16\textwidth]{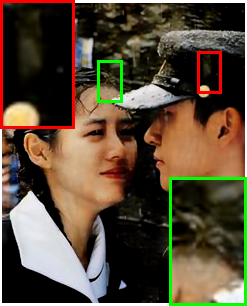}}
\vspace{-4mm}
\caption{Real rain streaks removal experiments under different scenarios. From left to right are input image, results of DSC\cite{luo2015removing}, LP \cite{li2016rain}, CNN \cite{fu2017removing}, DID-MDN\cite{derain_zhang_2018} and ours. Demarcated areas in each image are amplified at a 3 time larger scale.}
\label{real} 
\end{figure*}

\begin{figure*}[t]
\vspace{-3mm}
\centering
\includegraphics[height=3.5cm]{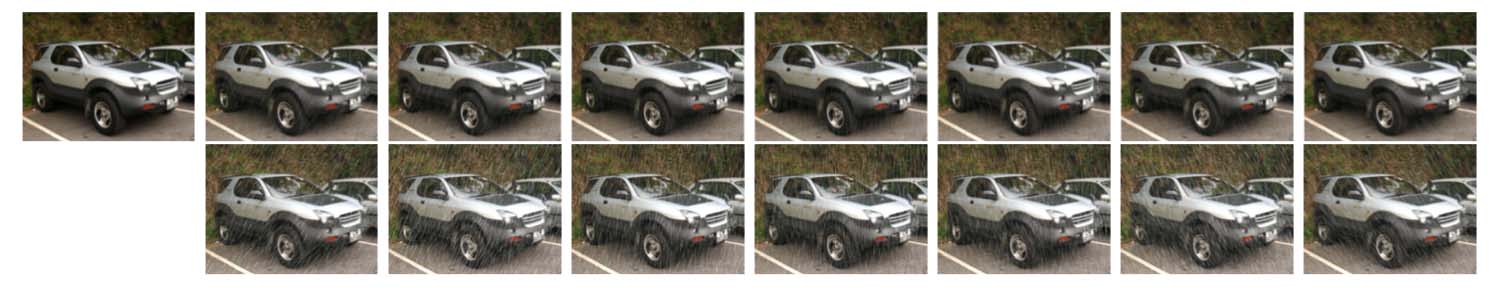}
\vspace{-8mm}
\caption{List of fourteen synthesized rain data types in our supervised data. The left image is the original one without rain streaks, and the right 14 ones are those superimposed with different rain types. The rain details can be more evidently observed by zooming in the images on a computer screen.}
\label{syn}
\vspace{-2mm}
\end{figure*}

\subsection{Experiments on real images}

The most direct way to evaluate a SIRR method is to see its visual effect of restoration results on the real world rainy images. We use the testing data selected from the Google search. To better represent the diversity of the real rain scenarios, we intentionally select images with different types of rain streaks as shown in Figure \ref{real}. 

To confirm the necessity of investigating transfer learning for this task, we list the complete synthesized rain types \cite{fu2017clearing} in our supervised training data in Figure \ref{syn}. The bias of rain between Figures \ref{real} and \ref{syn} is obvious and the transfer ability of our model can thus be substantiated. The visual effect of derained images verify that our method can remove more rain streaks and better keep the visual quality. Compared to other competing methods, our method can remove more amount of the rain streaks while still better keep the structure of image undamaged.

\section{Conclusion}
\label{sec5}

In this paper, we have attempted to solve the SIRR problem in a semi-supervised transfer learning manner. We train a CNN on both synthesized supervised and real unsupervised rainy images. In this manner, our method especially alleviates the hard-to-collect-training-sample and overfitting-to-training-sample issues existed in conventional deep learning methods designed for this task. The experiments implemented on synthesized and real images substantiate the effectiveness of the proposed method.

We admit that our model is still not almighty for all rainy image which could be extremely complicated to handle. The involvement of more elaborate priors on rain and background layers in training the network could be the future direction to further improve the performance for this task. Also this semi-supervised transfer learning methodology could be considered into other inverse problems as well. We wish to apply the human prior knowledge into the learning process of neural network framework, more sufficiently realizing the combination of data-based and model-based methods. The ultimate goal is to take advantage of both supervised data-based deep learning methods, which could shorten the testing time to fulfill the online requirement, and model-based method, to put the network training into a more explainable direction.
\vspace{-2mm}
\section*{Acknowledge}
\vspace{-2mm}
This research was supported by National Key R\&D Program of China (2018YFB1004300), China NSFC projects (61661166011, 11690011,61603292, 61721002, U1811461), National Science Foundation grant IIS-1619078, IIS-1815561, and the Army Research Ofice ARO W911NF-16-1-0138.

{\small
\bibliographystyle{ieee}
\bibliography{egbib}
}

\end{document}